\documentclass{article}

\usepackage{arxiv}
\usepackage[utf8]{inputenc}
\usepackage[T1]{fontenc}
\usepackage{amsfonts}
\usepackage{url}
\usepackage[natbibapa]{apacite}
\usepackage{doi}
\usepackage{bm}
\usepackage{graphicx}
\usepackage[font=small,labelfont=bf]{caption}
\usepackage{hyperref}
\usepackage[toc,page]{appendix}
\usepackage{booktabs}
\usepackage[table]{xcolor}
\usepackage{chngcntr}

\title{Creating Disasters: Recession Forecasting with GAN-generated Synthetic Time Series Data}

\author{Sam Dannels \thanks{University of Iowa, Department of Statistics and Actuarial Science. Email: sam.dannels@gmail.com}}

\date{Draft: February 2023}

\renewcommand{\headeright}{}
\renewcommand{\undertitle}{}
\renewcommand{\shorttitle}{Creating Disasters: Recession Forecasting with GAN-generated Synthetic Time Series Data}

\hypersetup{
pdftitle={Creating Disasters: Recession Forecasting with GAN-generated Synthetic Time Series Data},
pdfsubject={stat.AP},
pdfauthor={Sam ~Dannels},
pdfkeywords={},
}

\begin{document}
\maketitle

\begin{abstract}
A common problem when forecasting rare events, such as recessions, is limited data availability. Recent advancements in deep learning and generative adversarial networks (GANs) make it possible to produce high-fidelity synthetic data in large quantities. This paper uses a model called DoppelGANger, a GAN tailored to producing synthetic time series data, to generate synthetic Treasury yield time series and associated recession indicators. It is then shown that short-range forecasting performance for Treasury yields is improved for models trained on synthetic data relative to models trained only on real data. Finally, synthetic recession conditions are produced and used to train classification models to predict the probability of a future recession. It is shown that training models on synthetic recessions can improve a model's ability to predict future recessions over a model trained only on real data.
\end{abstract}

\section{Introduction}
One of the most difficult tasks in forecasting and predictive analytics is predicting a rare occurrence event. Events that occur with low frequency, and thus have low probability in the sample space, are difficult to adequately measure and incorporate into a model due to a lack of sufficient data. One such event is an economic recession. The National Bureau of Economic Research (NBER) tasks its Business Cycle Dating Committee with classifying business cycle peaks and troughs, which indicate the turning points between economic expansions and contractions. NBER classifies a recession as a time period that "involves a significant decline in economic activity that is spread across the economy and lasts more than a few months" \citeyearpar{NBERweb}. Between January 1965 and January 2023, NBER lists only eight recessions. 

Perhaps the most famous leading indicator of a recession is an inversion of the yield curve. The yield curve measures how yields vary for Treasury bills and bonds based on their maturity. Generally, short-term bills will have lower yield than longer-term bonds. Occasionally, the yield curve will "invert", meaning that market forces have driven short-term yields above long-term yields. It has been well documented that this phenomenon can be a useful indicator that a recession is imminent \citep{Kessel, Estrella&Hard, Bernard}, and that utilizing the spread between short and long-term yields can predict recessions better than many other leading indicators \citep{NYFed}. 

While yield curve inversions may show promise in predicting recessions, difficulty arises in training and testing a model on the few available recessions for which there is adequate data. In order to better train a model to predict recessions, one would need to observe more recessions. Recent advances in the ability of generative adversarial networks (GANs) to create synthetic replicates of time series data may provide a solution to this problem. The goal of this paper is to demonstrate that generating synthetic replicates of Treasury bond yield data shows initial promise in improving short-term forecasts of yield movements, as well as strengthens the signal provided by recession probability models making use of the Treasury yield spread.

\section{Background}
\subsection{Synthetic Data}
Synthetic data is an artificial replicate of a real data example that is ideally drawn from the same distribution as the original example. Synthetic data can be generated manually by experts or through a statistical or machine learning algorithm. The resulting synthetic data should have high fidelity, that is it should maintain the statistical properties of the original example it is based on while not being an exact replicate of the original data \citep{Nikolenko, DoppelGANger}. %In other words, synthetic data should have all of the same properties and contain much of the same information as the original data set it is meant to replicate, but is not merely a copy of the original data.

Synthetic data is useful for the purpose of sharing data for collaboration across organizations without revealing any proprietary information \citep{DoppelGANger}, providing more samples or increasing the frequency of novel samples in a sparse data set \citep{Nikolenko}, and in some cases can increase predictive performance over models trained on traditional data \citep{liver, DTW_augment}.

There are many means of producing synthetic data, but the recent success of GANs in producing high-fidelity replicates of images \citep{cyclegan, stylegan} has led to their popularity among researchers. One particular GAN that has demonstrated success in producing synthetic samples of multidimensional time series data is called DoppelGANger \citep{DoppelGANger}. In this paper, DoppelGANger will be the model used to produce high-fidelity replicates of 1-year and 10-year Treasury bond yields. Before moving on to the specific architecture of DoppelGANger, we will review the general structure and training of GANs.

\subsection{GANs}
GANs were first introduced by Goodfellow et al. in the 2014 work "Generative Adversarial Networks". The following section is a summary of the ideas proposed in that paper. The basic idea behind the method they introduced is that there are two models - a generative model, \emph{G}, and a discriminative model, \emph{D} - that compete in a minimax game. The generative model (generator) is given latent variables (noise) as inputs and attempts to produce a mapping from the latent space to some desired data space. An example would be a generator that produces realistic images of cats. It would attempt to learn a mapping from the latent space to a complex probability space that represents realistic-looking photos of cats. The discriminative model (discriminator) is given inputs from both the real data and the fake samples produced by the generator and tasked with distinguishing between the two. It produces a single probability that the input came from the data rather than the generator. In our example, the discriminator would provide a probability that the given input is a real image of a cat and thus not a computer-generated image of a cat. The two models are trained in an adversarial way, such that the generator seeks to trick the discriminator with its fake samples while the discriminator seeks to correctly classify each input it is given as real or generated data. The ultimate goal is that this minimax optimization training scheme will result in a generator that is able to produce high-fidelity samples from the same space as the original data.

More specifically, allow $\bm{x}$ to be the data samples for training the GAN and $\bm{z}$ to be the latent variables provided to the generator. Then the minimax problem can be represented by:

$$\min_G \max_D \mathop{\mathbb{E}}_{\bm{x} \sim p_{data}} [log D(\bm{x})] + \mathop{\mathbb{E}}_{\bm{z} \sim p_{z}} [log(1-D(G(\bm{z})))]$$

While \emph{G} and \emph{D} can be a wide variety of differentiable functions, in practice they are most commonly deep neural networks. GANs require few assumptions about the data space and can be trained through common backpropagation algorithms, which avoid the need for difficult parametric likelihood maximization and Markov Chains \citep{DoppelGANger, GAN_intro}. In practice, \emph{G} and \emph{D} are trained iteratively. \emph{D} is trained for $k$ steps while \emph{G} is held fixed, and then \emph{G} is trained for a single step while \emph{D} is held fixed. When updating \emph{G}, the gradient is computed for both \emph{D} and \emph{G}, because the gradient backpropagation must pass through \emph{D}, but the weights are only updated for \emph{G}. $k$ is a hyperparameter whose optimal value can vary based on the task at hand \citep{GAN_intro, Glassner}.

When training GANs, there are a few extra considerations to improve the chances of convergence. Dropout, the practice of regularizing a neural network by stochastically dropping nodes throughout the training process, can help avoid overfitting \citep{Goodfellow-et-al-2016}. To avoid overfitting GANs, it is important to include dropout in the training process, especially in the discriminator \citep{Goodfellow-et-al-2016}. It has also been shown that in practice, training \emph{G} to maximize $log D(G(\bm{z}))$ rather than minimize $log(1-D(G(\bm{z})))$ can improve the efficiency of training \citep{GAN_intro}. 

GAN training can be very unstable. GANs are especially prone to "mode collapse", where many inputs $\bm{z}$ produce the same output \citep{GAN_intro}. In our cat example, the goal is to train a generator that can create many different images of photorealistic cats. While training, the generator is able to fool the discriminator with an image of a black cat. Rather than move to other areas of the data's probability space, the generator remains in the area that produced the black cat and continues to produce the same image of a black cat over and over. The generator is technically meeting its objective of fooling the discriminator, but there is no diversity in the resulting generated images. For these reasons, training GANs requires a lot of experimentation and experience. While GANs can be tricky to train for the reasons stated above, when they are properly trained, they can produce very useful results.

\subsection{DoppelGANger}
DoppelGANger \citep{DoppelGANger} is a GAN designed to generate multivariate time series and corresponding metadata. Metadata are fixed attributes of the data that are not a part of the time series, but still contain relevant information about the series. For example, a data set may contain 4 time series, each corresponding to the change of a certain stock's price over time. Relevant metadata may be a ticker symbol label for each stock, the industry each stock belongs to, or any other fixed attribute. DoppelGANger was designed with the intent of producing synthetic networked systems data for data sharing, but works as a general framework for producing many types of synthetic time series data. The design of DoppelGANger alters the architecture of traditional GANs in a few crucial ways to tackle problems in a time series setting.

The first alteration has to do with the type of neural network used in the GAN. DoppelGANger, and other time series GANs \citep{Arnelid, Mogren}, rely on recurrent neural networks (RNNs) that are able to capture temporal correlations. More specifically, DoppelGANger uses a popular type of RNN called long short-term memory (LSTM). To illustrate the basic idea of RNN generators, say there is a univariate sequence of length $\tau$. At step $i$, the RNN will generate a value one step ahead, $x_{i+1}$, using information from previous values in the sequence and the current state, $\{x_1, x_2, \ldots, x_i\}$. At each step the RNN keeps an encoded memory that summarizes relevant information about the previous states, $\bm{f}(\{x_1, x_2, \ldots, x_{i-1}\}; \bm{\theta}$), allowing it to track temporal patterns \citep{Goodfellow-et-al-2016}. The RNN passes over the sequence $\tau$ times to produce the entire series. This idea can be extended to multivariate time series. See Appendix \ref{appendix:A} for more details on RNNs and LSTM. A common problem with RNN generators is that long sequences require a very large number of passes of the RNN, which make it difficult to capture longer-term correlations \citep{DoppelGANger}. To handle this problem, the creators of DoppelGANger use what they call "batch generation". Instead of generating one step ahead in each pass, DoppelGANger generates $S$ steps ahead, which reduces the necessary number of passes. The authors of DoppelGANger suggest a value of $S$ such that $\tau / S$ is around 50, but $S$ can be tuned for specific tasks.

The next aspect of DoppelGANger is designed to mitigate mode collapse. The model uses a unique kind of normalization for time series exhibiting a wide range of values across samples and series. For example, say in a given sample one series has a range of 0 to 100, while another series in the sample has a range of -5 to 5. Rather than normalize by the global minimum and maximum (-5 and 100) for all series, each series is normalized by its individual minimum and maximum in that sample, and these values are stored as metadata for that series in the form $\frac{min+max}{2}$. DoppelGANger learns these metadata and generates a new minimum and maximum for rescaling each generated series. Experimentation indicates that this can help reduce mode collapse in many cases where ranges greatly vary between series \citep{DoppelGANger}.

\begin{figure}
    \centering
    \includegraphics[scale = .7]{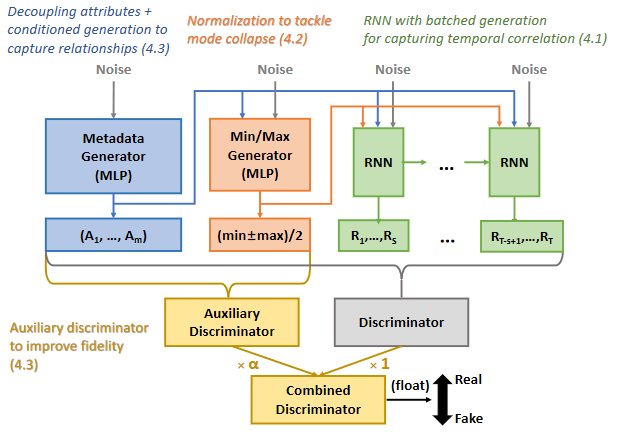}
    \captionsetup{width=.75\linewidth}
    \caption{
    DoppelGANger's Architecture \\ 
    Source: "Using GANs for Sharing Networked Time Series Data: Challenges, Initial Promise, and Open Questions" by Lin et al.
    \href{https://dl.acm.org/doi/pdf/10.1145/3419394.3423643}{URL: https://dl.acm.org/doi/pdf/10.1145/3419394.3423643} \\
    \href{https://creativecommons.org/licenses/by/4.0/}{Copyright Information: Creative Commons Attribution 4.0 International Public License}
    }
    \label{fig:fig1}
\end{figure}

A useful feature of DoppelGANger is that it does not only generate synthetic time series, but also jointly generates synthetic attributes (metadata) for those series. This is accomplished by using multiple generators. The first generates metadata using a standard multilayer perceptron (MLP). These generated metadata are then given as inputs to another MLP that generates the minimum and maximum metadata used for the normalization scheme described in the paragraph above. Then all of the combined metadata are given as inputs to the LSTM that generates the synthetic time series. To account for the different dimensionality and types of data being generated, there is an additional "auxiliary discriminator", $D_{aux}$, that measures loss on only the metadata. The Wasserstein loss of the two discriminators are combined by the weighting scheme: $\min_G \max_{D, D_{aux}} \mathcal{L}(G, D) + \alpha\mathcal{L}_{aux}(G, D_{aux})$. A diagram of DoppelGANger's architecture can be seen in Figure \ref{fig:fig1}.

\section{Synthetic Data Generation}
\subsection{Data}
\begin{figure}
    \centering
    \includegraphics[scale = .45]{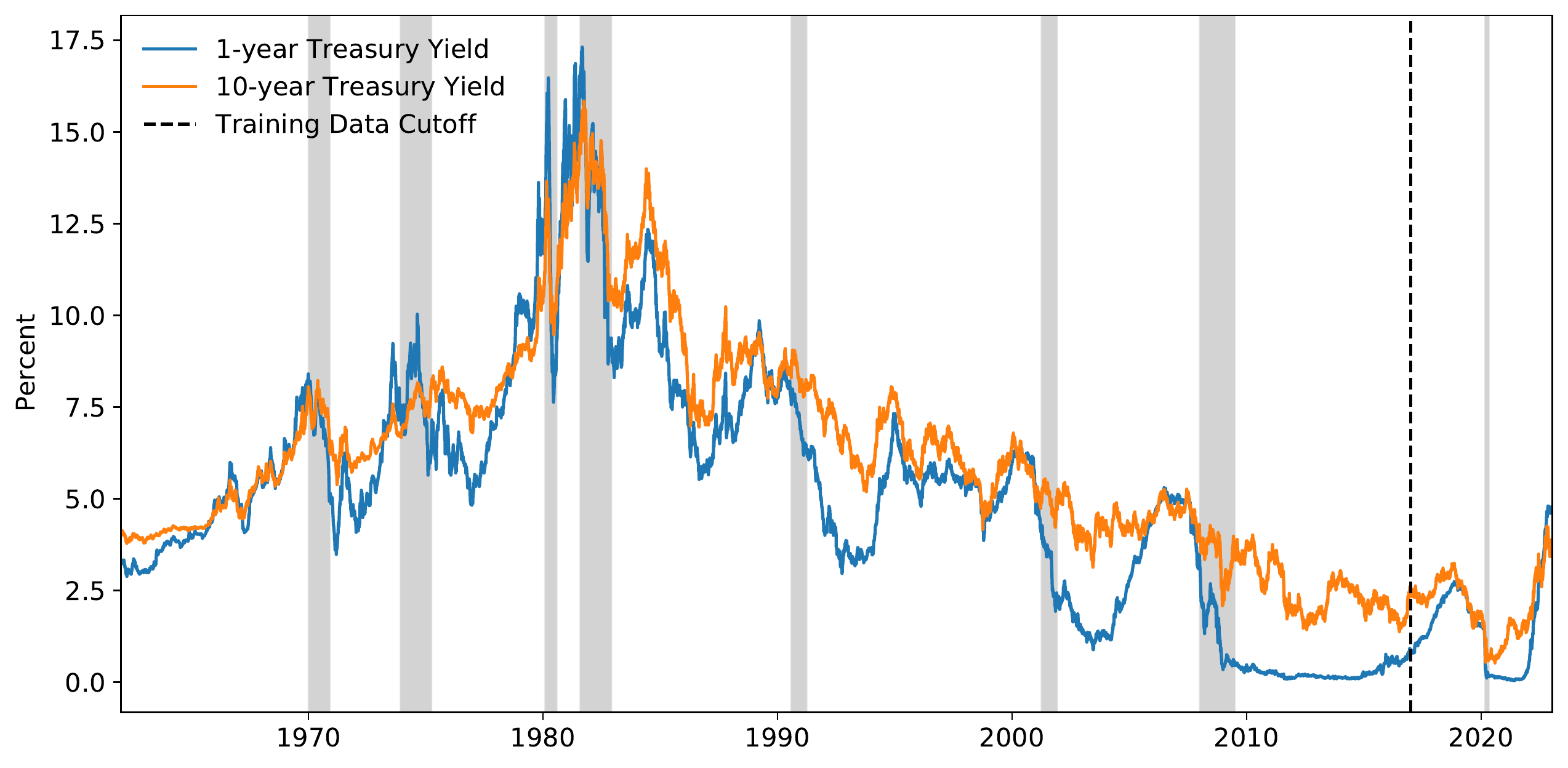}
    \captionsetup{width=.75\linewidth}
    \caption{
    U.S. Treasuries Yield Data\\
    Note: Grey areas indicate recession \\
    Sources: Board of Governors of the Federal Reserve System, Federal Reserve Bank of St. Louis, NBER
    }
    \label{fig:fig2}
\end{figure}

Data were collected from the Federal Reserve Economic Data (FRED) website hosted by the Federal Reserve Bank of St. Louis. Data were collected at a daily frequency for market yield on U.S. Treasuries at both a 1-year and 10-year constant maturity \citep{1year_data, 10year_data}. There are many possible variations of short and long-term yield spread, but for this study, 1 and 10-year maturities were selected because they have long histories available. An indicator for recession according to NBER was also collected at a daily frequency \citep{recess_data}. Note that yield data is only collected on weekdays. Figure \ref{fig:fig2} shows the Treasury yield data. For the first task, DoppelGANger was trained on data ranging from 01-02-1962 to 12-30-2016. The remaining data (from 01-03-2017 to 01-23-2023) was not used in training so that forecasting performance could be evaluated on that time period.

\subsection{Training}

As with all deep learning models, DoppelGANger requires many samples to train on. So the training set was split into samples of 125 day segments to be used as inputs to DoppelGANger. The features used in training were 1-year Treasury Bill yield and 10-year Treasury Bond yield. For metadata, a single attribute was added to each sample indicating whether or not a recession occurred at any point in that particular 125 day span.

A PyTorch implementation of DoppelGANger created by Gretel \citep{gretel.ai} was used in training. Samples were 125 days in length and after some experimentation, $S$ was set to 5; meaning there were 25 RNN passes over each sample. The normalization scheme using the local minimum and maximum was applied to the yield data. In neural network training, the learning rate is a hyperparameter that regulates the magnitude of updates at each step in the training process. Smaller learning rates are recommended for training and ensure that the gradient descent algorithm does not take too large of steps and overshoot minimums \citep{Glassner}. The generator and discriminator learning rates were both set equal to $10^{-4}$, and training was run for 2000 epochs.

\subsection{Evaluating generated data}

\begin{figure}
    \hspace*{-1cm}
    \centering
    \includegraphics[scale = .47]{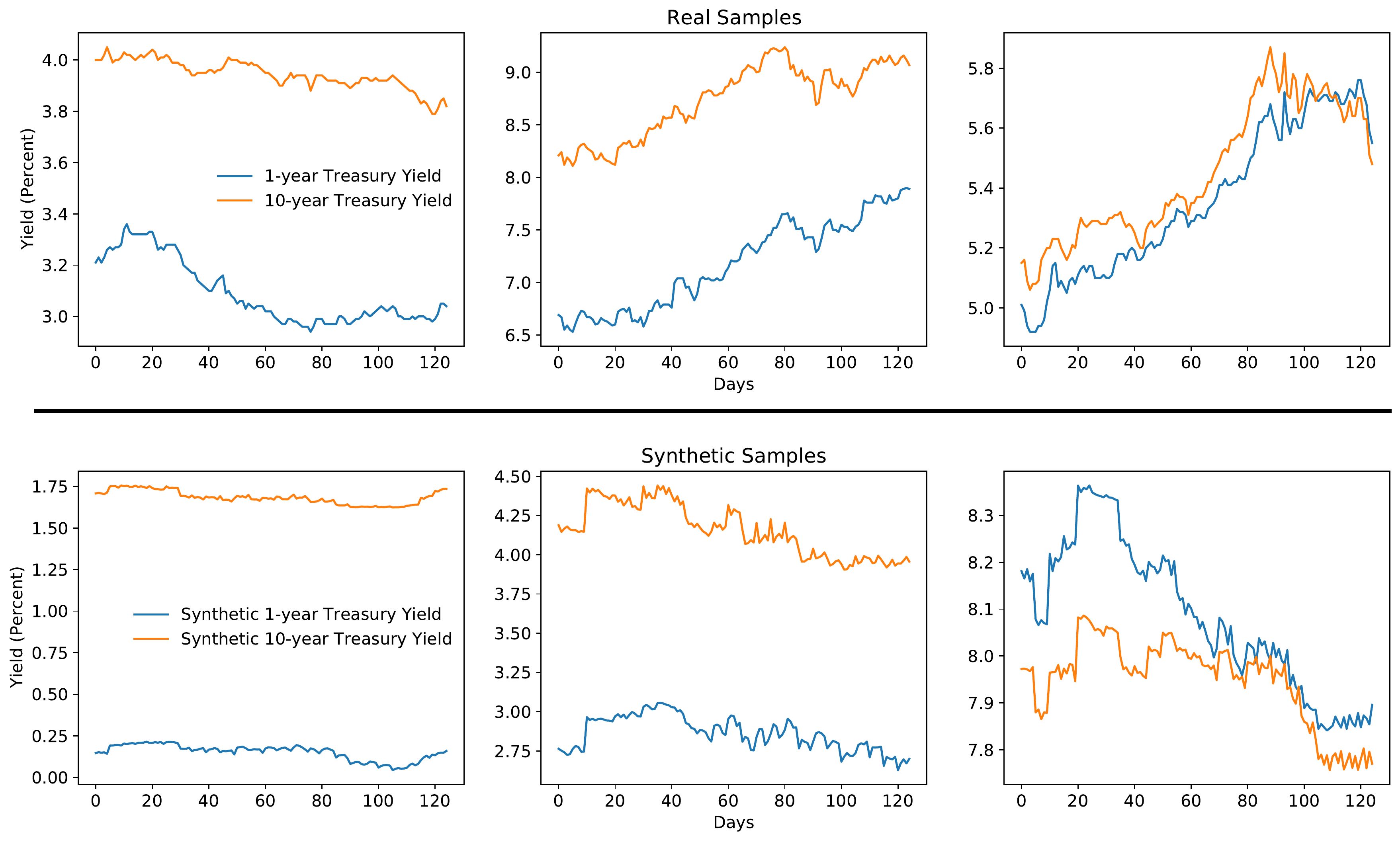}
    \captionsetup{width=.75\linewidth}
    \caption{
    Comparison of real samples to synthetic samples\\
    Note: The top row contains true data samples from the training set.  The bottom row contains generated data from DoppelGANger. The bottom right is an example of a generated yield curve inversion. \\
    Sources: Board of Governors of the Federal Reserve System, Federal Reserve Bank of St. Louis, NBER, author
    }
    \label{fig:fig3}
\end{figure}

After training the model, 1000 synthetic samples were generated. These are each 125 days in length and paired with recession indicator metadata. Figure \ref{fig:fig3} shows three real samples from the training data and three synthetic samples. Evaluating the data qualitatively, the synthetic samples appear quite similar to what one might expect from the true data. The bottom right panel of Figure \ref{fig:fig3} is an example of a generated yield curve inversion, indicating that DoppelGANger was able to learn from those unique cases as well as the standard cases where the 10-year yield should be significantly higher than the 1-year yield. As far as metadata generation, approximately $18\%$ of training samples contained a positive indicator for recession. While approximately $17\%$ of generated samples contained a positive indicator for recession, meaning the samples containing recession data remained roughly proportional to the true data. The key is that the trained generator now has the ability to produce as many fake recessions as needed.

\begin{figure}
    \centering
    \includegraphics[scale = .4]{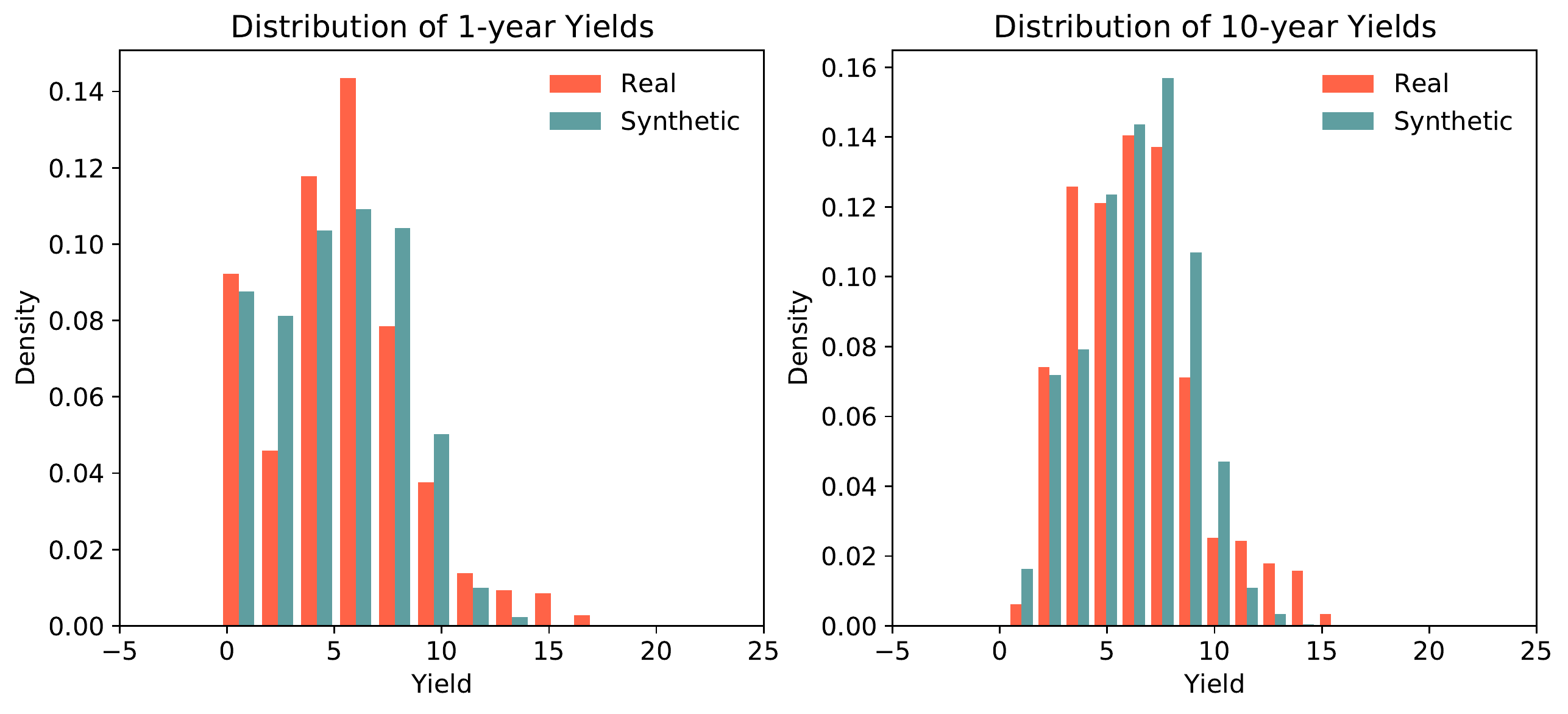}
    \captionsetup{width=.75\linewidth}
    \caption{
    Distributions of real and generated yields\\
    Source: author's calculations
    }
    \label{fig:fig4}
\end{figure}

In the training set, the correlation between 1-year and 10-year Treasury bond yields was 0.946. In the generated samples, the correlation was 0.945. Figure \ref{fig:fig4} shows histograms for the 1-year and 10-year training data against their synthetic counterparts. The synthetic data approximately follows the same distribution as the training data with a bit more noise. In rare cases, the synthetic data may produce a negative yield or an outlier in terms of magnitude. These are not ideal, but given that they are a negligible proportion of the generated samples, they likely do not have much affect on downstream predictive tasks.

Given that the training data is split into segments, the model appears to struggle to capture the strength of the long-term autocorrelations. For both 1-year and 10-year yields, the autocorrelation remains high even at a 100 day lag when estimated on the full range of training data. It is difficult to assess autocorrelation for separated samples, however. The method used was to assess the autocorrelation within each 125 day sample and average the autocorrelations across all samples. This allows for comparison with the generated samples, which are independent 125 day segments. 

\begin{figure}
    \centering
    \includegraphics[scale = .4]{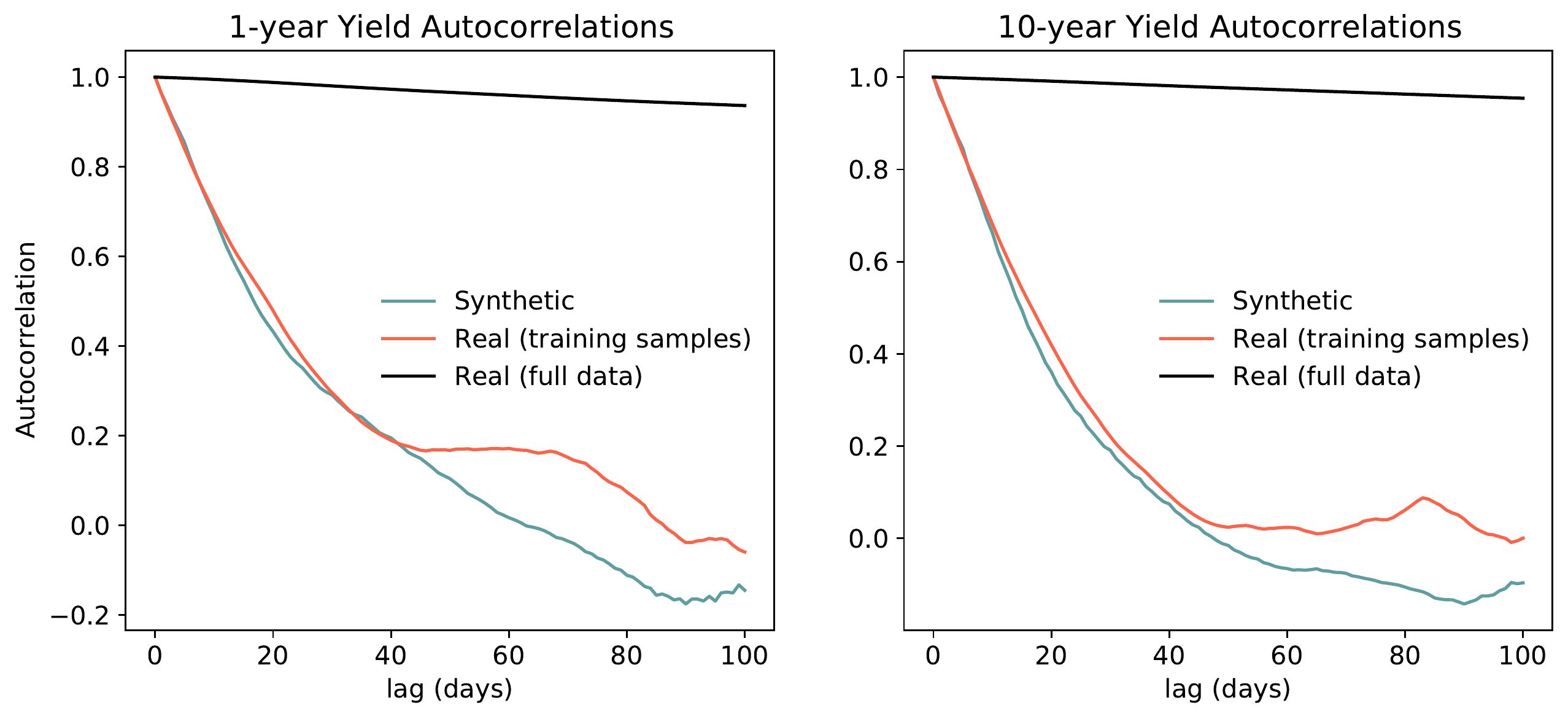}
    \captionsetup{width=.75\linewidth}
    \caption{
    Estimated autocorrelations \\
    Note: The synthetic and real (training samples) values are estimated as an average of autocorrelations across many samples. The real (full data) autocorrelation is estimated on the full range of training data before being split into training samples. \\ 
    Source: author's calculations
    }
    \label{fig:fig5}
\end{figure}

When using this method, the training samples and generated samples show fairly similar patterns in their autocorrelations; however, they both show more rapid decline in autocorrelation than the estimates from the full training data. You can see these patterns in Figure \ref{fig:fig5}. So it appears that DoppelGANger was able to replicate the autocorrelations from the training samples, but those samples underestimate the extent to which autocorrelation remains high in the yield curve data. It should also be noted that higher lags have fewer data to estimate autocorrelation, and thus had very high variance across samples. It is possible that lengthening the training samples to include more days could help improve replication of long-term autocorrelations, but would come at the cost of fewer training samples. There is a trade-off to consider between length of samples and quantity of samples when preparing training data. This should be considered when deciding the horizon of forecasts that will be produced in downstream tasks.

The results shown in this section are an illustrative example of a successful training of DoppelGANger to Treasury data. Not every training attempt was successful, however. Getting useful synthetic data required a bit of experimentation and tuning, and even with the same training scheme, results can vary widely. In several training attempts, mode collapse resulted in nearly all generated samples containing perfectly straight lines. DoppelGANger, like most GANs, can be unstable and difficult to train but produces great results when properly trained. 

\section{Forecasting Treasury Yields}
The first task is to evaluate forecasting performance of models trained on synthetic data against those trained on real data. In order to do that, an LSTM model was trained to produce 1-day ahead forecasts of 1-year and 10-year Treasury yields based on the previous 25 days of data.

Using actual data for 1-year and 10-year Treasury yields as shown in Figure \ref{fig:fig2}, training samples were created by splitting the data into a rolling window of 25 days. That is, the first 25 days of data make up the first sample, with day 26 being its associated response. Then the window is moved forward by one day. Days 2-26 make up the second sample, with an associated response from day 27, and so on. These samples were created over the time frame 01-02-1962 to 12-30-2016, the same range of data that trained DoppelGANger. This produced 13,710 samples and associated one-day ahead responses. We will refer to this set of samples as the real data training set.

As seen in section 3, DoppelGANger produced 1000 synthetic samples of yield data. These synthetic samples were used to create forecast training data in the same fashion as the real data training set described above. A rolling 25 day window split up training samples and responses within each synthetic sample. Each 125 day segment of synthetic data was able to generate 100 training samples and responses for a total of 100,000 synthetic training samples. We will refer to this as the synthetic data training set. 

Finally, a training set was created by combining the real data training set and the synthetic data training set. The idea is that the LSTM model will be trained on real data that is augmented by synthetic data in order to allow for more samples. This training set has 113,710 samples and will be referred to as the combined data training set. 

The model trained for forecasting is a deep neural network with two stacked LSTM layers, each with a hyperbolic tangent (tanh) activation function, followed by a dropout layer and a dense layer with 2 nodes for the 2 features (1-year and 10-year Treasury yield). A diagram of this model can be seen in Figure \ref{fig:figC1}. The model used a mean-squared error loss function and Adam optimizer \citep{Adam}. This model was trained with each of the three training sets - real, synthetic, and combined - each for 50 epochs. Each trained model then produced one-day ahead forecasts for a test set containing real Treasury yield data from 01-03-2017 to 01-11-2023. Forecasting performance over this test set, measured by root mean square error (RMSE) and mean absolute percentage error (MAPE), can be seen in Table \ref{tabl:tabl1}.

\begin{table}
    \centering
    \caption{Forecasting performance for 1-day ahead forecasts over period January 3, 2017 to January 11, 2023.}
    \begin{tabular}{lrrr}
        &&& \\
        & Real & Synthetic & Combined \\
        \midrule
        1-year Treasury RMSE & 0.077  & \cellcolor{red!25} 0.046 & 0.072 \\
        \midrule
        1-year Treasury MAPE & \cellcolor{red!25} 7.793 & 9.413 & 8.368 \\
        \midrule
        10-year Treasury RMSE & 0.253 & \cellcolor{red!25} 0.062 & 0.073 \\
        \midrule
        10-year Treasury MAPE & 15.903 & \cellcolor{red!25} 2.709 & 3.484 \\
        \bottomrule
    \end{tabular}
    \label{tabl:tabl1}
\end{table}

\begin{table}
    \centering
    \caption{Forecasting performance for 15-day ahead forecasts over period January 3, 2017 to January 11, 2023.}
    \begin{tabular}{lrrr}
        &&& \\
        & Real & Synthetic & Combined \\
        \midrule
        1-year Treasury RMSE & \cellcolor{red!25} 0.244 & 0.355 & 0.307 \\
        \midrule
        1-year Treasury MAPE & 77.736 & 84.610 & \cellcolor{red!25} 58.448 \\
        \midrule
        10-year Treasury RMSE & 0.425 & 0.543 & \cellcolor{red!25} 0.269 \\
        \midrule
        10-year Treasury MAPE & 25.075 & 30.503 & \cellcolor{red!25} 13.557 \\
        \bottomrule
    \end{tabular}
    \label{tabl:tabl2}
\end{table}

\begin{figure}
    %\hspace*{-1cm} 
    \includegraphics[scale = .4]{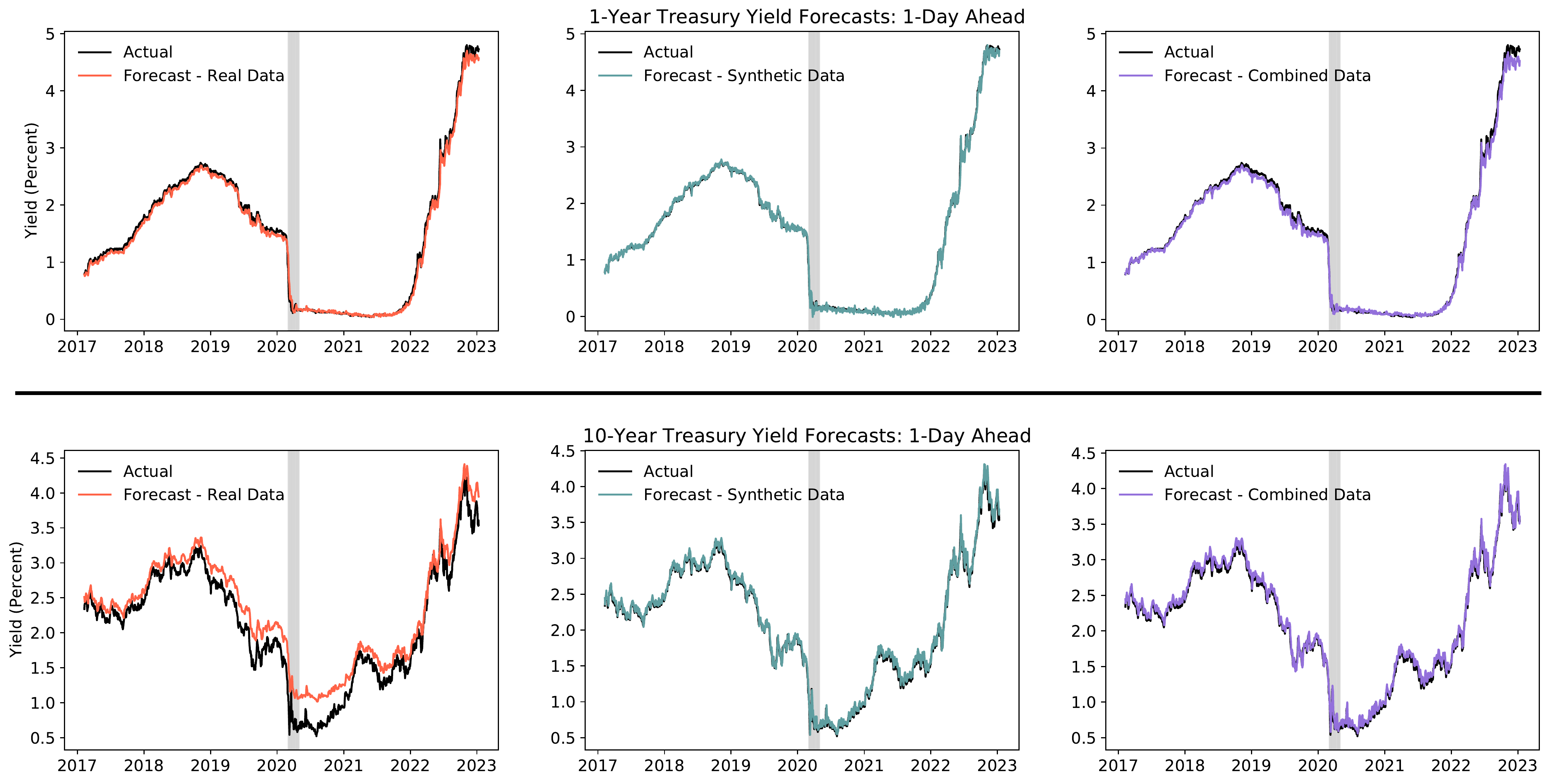}
    \captionsetup{width=.75\linewidth}
    \caption{
    1-day ahead forecasts for each type of training data\\
    Sources: Board of Governors of the Federal Reserve System, Federal Reserve Bank of St. Louis, NBER, author
    }
    \label{fig:fig6}
\end{figure}

The LSTM model trained on synthetic data outperformed the other models in every metric except for the MAPE for 1-year Treasury yield. Interestingly, the model trained on combined data performed much better than the model trained on real data for the 10-year Treasury yield, but still did not match the performance of the model trained on synthetic data only. These results imply that models trained on synthetic data can improve forecasting performance over a model trained on only real data, likely due to the increased availability of training data. 

The same LSTM model architecture was then used again to test the forecasting performance of longer-range forecasts. This time, rather than providing responses of a single day ahead, the response was a vector of 15 days beyond the end of the training sample. The LSTM was thus trained to take a sample of 25 days and predict yield values for the following 15 days. Again, three models were trained on real data, synthetic data, and combined real and synthetic data. Forecasting performance was evaluated for the $15^{th}$ day of the generated predictions, and these metrics can be found in Table \ref{tabl:tabl2}.

\begin{figure}
    %\hspace*{-2cm} 
    \includegraphics[scale = .4]{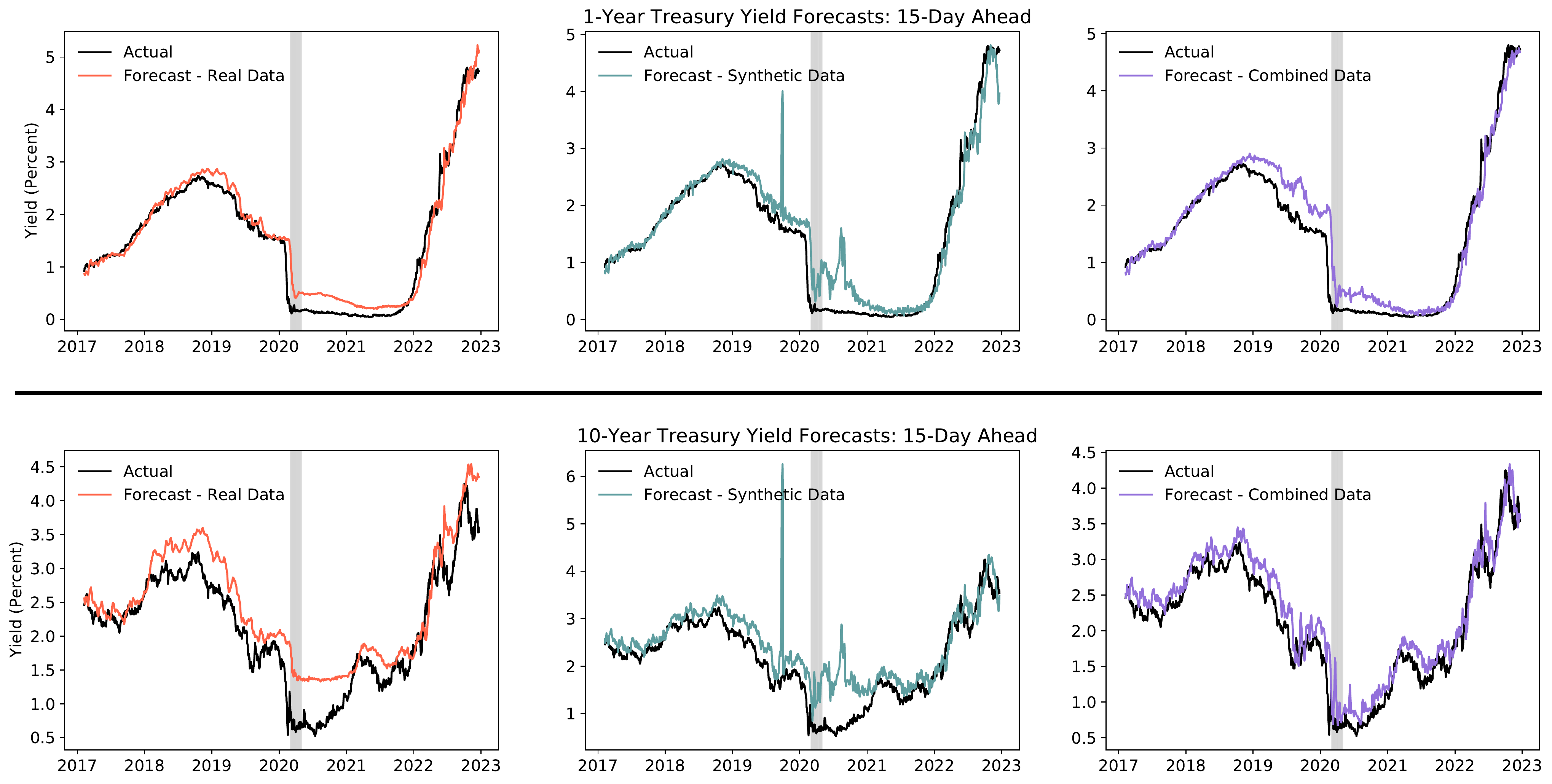}
    \captionsetup{width=.75\linewidth}
    \caption{
    15-day ahead forecasts for each type of training data\\
    Sources: Board of Governors of the Federal Reserve System, Federal Reserve Bank of St. Louis, NBER, author
    }
    \label{fig:fig7}
\end{figure}

The longer-range forecasting model trained on only synthetic data consistently had the worst performance. This is not surprising, as it was shown in section 3 that the synthetic data struggled to capture the longer-range autocorrelation of the true data. In longer-range forecasting, the model trained on combined real and synthetic data fared the best. This could potentially be due to the fact that the combined data model benefits from both learning the true structure of the real data and an increased availability of training samples from the synthetic data. Overall, it appears there is a benefit to using synthetic data in forecasting Treasury yields. 

\section{Predicting Recessions}
The second task of this study is to test if using synthetic data can improve the prediction of an upcoming recession. The framework for predicting an upcoming recession is to train a classification model to provide a probability that a recession will occur any time in the next 250 days. The challenge for these models is that when using available historic data, there are very few actual recessions for training and testing. In the range of dates for which Treasury yield data is available, there are only 8 recessions that have occurred. One of those recessions (2020) is a result of COVID-19, and could not possibly be predicted by economic factors. That leaves 7 recessions to train and test a model; however, with synthetic data, we are able to produce as many fake recessions as needed. 

\subsection{Generating data}
In order to build classification models to predict upcoming recessions, new synthetic data must be generated. The primary changes in this round of synthetic data generation is that a new attribute (metadata) was added to indicate a future recession and DoppelGANger was trained on a smaller set of data. The samples are also only 30 days long, rather than the 125 day samples created before.

For this task, DoppelGANger was trained on the time period 01-02-1962 to 12-31-1984. This time period contains 4 recessions, which leaves 3 recessions for testing downstream predictive performance, as seen in Figure \ref{fig:figB1}. In addition to the indicator for a recession used in the previous training of DoppelGANger, an indicator for a future recession was added. This future recession indicator takes value 1 for samples in which a recession occurred any time in the 250 days following the sample, and 0 otherwise. In other words, this indicator labels each sample by whether or not it precedes a recession by at most 250 days. The features are still 1-year and 10-year Treasury yields, while the metadata now has two dimensions - recession indicator and future recession indicator. 

The generated samples from this version of DoppelGANger performed equally well in terms of fidelity metrics as the generated samples displayed in section 3. This iteration of DoppelGANger was also able to capture the correlation between features and the overall distributions of both 1-year and 10-year Treasury yields, but similarly struggled to capture the autocorrelation structure. Full metrics can be found in Appendix \ref{appendix:B}.

\subsection{Classification models}
To test if synthetic data can improve a model's ability to predict future recessions, two models were used - logistic regression and an LSTM classifier. The goal is to give these models the previous 30 days of 1-year and 10-year Treasury data as an input and produce a predicted probability that a recession will occur at some point in the following 250 days. Each sample of 30 days in the training data is paired with an indicator for whether or not a recession occurred in the 250 days following that sample, including the synthetic data which generated that future recession indicator as metadata.

The logistic regression specification was treated like an autoregressive model in that each day in the 30 day sample was input as its own feature. In other words, each training sample had 60 features - 1-year and 10-year Treasury yields for the most recent day and 29 lags for each variable. In order to control for a large amount of features, the models were fit with an $L_1$ regularization. This means that there was a penalty on the estimated coefficients which shrinks them toward zero and performs an automated form of feature selection to reduce the dimensionality of the model. The optimization problem for logistic regression under $L_1$ regularization is \citep{logistic}:

$$\min_{\bm{\beta}} \sum_{i=1}^M -log \ p(y^{(i)} | \bm{x}^{(i)} ; \bm{\beta}) + \lambda ||\bm{\beta}||_1$$

Where $\bm{\beta}$ are the estimated coefficients and $\lambda$ is a hyperparameter that controls the amount of regularization. The logistic regression uses a gradient descent algorithm to minimize the negative log-likelihood of the model with the added constraint that the $L_1$ norm of the model be kept reasonably small.

The second model used was an LSTM-based classification model. It works in much the same way as the LSTM models used in the forecasting task of section 4, with the exception that it produces probabilities of recession rather than forecasting. The architecture of the LSTM classification model contained a single LSTM layer with a hyperbolic tangent activation function followed by a dropout layer, then a dense layer with 100 nodes and a hyperbolic tangent activation function. Finally, there was a dense layer with 2 nodes and a softmax activation to ensure that the model produced probabilities. This model was trained with a categorical cross-entropy loss function and an Adam optimizer for 50 epochs. A diagram of this model can be found in Figure \ref{fig:figC2}. 

Both the logistic regression and LSTM-based classifier were trained on samples of real data, synthetic data, and a combination of real and synthetic data. The real data comprised 30 day rolling window samples from 01-02-1962 to 12-31-1984, the same range of dates that trained DoppelGANger to produce synthetic data for the classification task. Thus, there were 5,701 real training samples. The synthetic training data comprised 50,000 30-day samples produced by DoppelGANger and their associated future recession indicator. The combined training data simply combined both the real and synthetic training samples into one training set.

The output of each model is a probability of recession in the next 250 days based on the previous 30 days of 1-year and 10-year Treasury yield data.

\subsection{Evaluating Recession Predictions}
The test set to evaluate each model's ability to predict future recessions covered the time period 01-02-1985 to 06-30-2009. This time period contains 3 recessions, and thus provides the opportunity to see how strong of a signal each model produces leading up to a new set of recessions. This section attempts to evaluate the performance of each model at assigning high probabilities of recession in the days leading up to a recession and low probabilities when a recession is not imminent. 

It should be noted that performance in this setting is quite difficult to evaluate due to the somewhat subjective nature of the classification labels. It is somewhat arbitrary to say a recession signal is only good if it appears within 250 days of the recession. So while strong recession signals more than 250 days out are considered false positives for this paper's purposes, others may argue those are valid early signals. The other aspect to consider is that policy actions are often taken to prevent recessions that are not considered by these models. So while a high probability of recession may seem like a false positive, it is possible that the model was signaling a recession that may have occurred in the absence of policy action. With these caveats in mind, we will attempt to evaluate each model as objectively as possible. 

\subsubsection{Logistic Regression Performance}
\begin{figure}
    %\hspace*{-2cm}
    \centering
    \includegraphics[scale = .45]{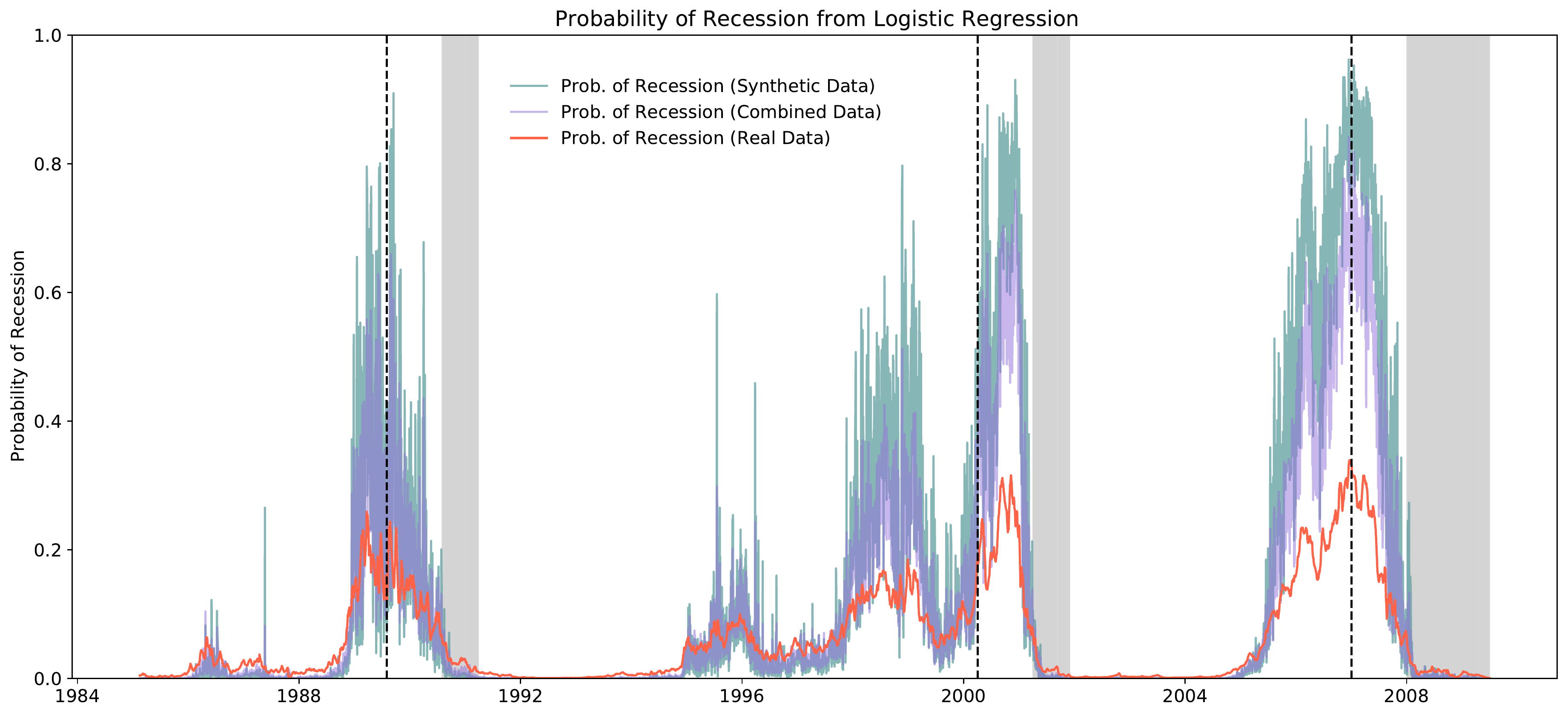}
    \captionsetup{width=.75\linewidth}
    \caption{
    Probability of Recession as predicted by logistic regression \\
    Note: Grey areas indicate recession. Black dashed lines indicate 250 days prior to the start of the next recession. \\
    Source: NBER, author's calculations
    }
    \label{fig:fig8}
\end{figure}

The probability of recession predictions produced by each logistic regression model can be found in Figure \ref{fig:fig8}. The model trained on only real data gives very weak signals, never producing a probability above 30 percent. The model trained on synthetic data follows much the same pattern as that of the model trained on real data, but produces much stronger signals, often reaching probabilities above 80 percent. The downside is that this model is much noisier. This is not surprising, as the synthetic data inherently contains more noise than the real data. The model trained on the combined data is similar to the model trained on synthetic data, though with a slightly weaker signal. It should be noted that the models that trained on additional synthetic recessions were better able to recognize upcoming recessions, but were also more sensitive to false positives. This can be seen in the period preceding the 2001 recession. The models trained with synthetic data gave very strong recessionary signals in 1996 and again in 1998-1999. 

The ROC curves (a graph that plots the true positive rate against the false positive rate) for each logistic regression model can be seen in Figure \ref{fig:fig10}. A useful summary of classification performance is the area under the ROC curve (AUC). The higher the AUC, the better the predictive performance. In the case of logistic regression, the ROC curves are very similar. The model trained on real data has a slight edge over the other two. This follows intuitively from the fact that all three models tend to give the same signal, albeit at very different levels. Rescaling the real data model such that a 25 percent probability of recession is considered a strong indicator of an upcoming recession puts it in line with the other models, with the added benefit of less noise. So while synthetic data does not appear to improve performance for the logistic regression, it at least produces estimates that are comparable to a model trained on real data - a good sign that DoppelGANger has produced high-fidelity synthetic data.

\subsubsection{LSTM Classifier Performance}
\begin{figure}
    %\hspace*{-2cm}
    \centering
    \includegraphics[scale = .45]{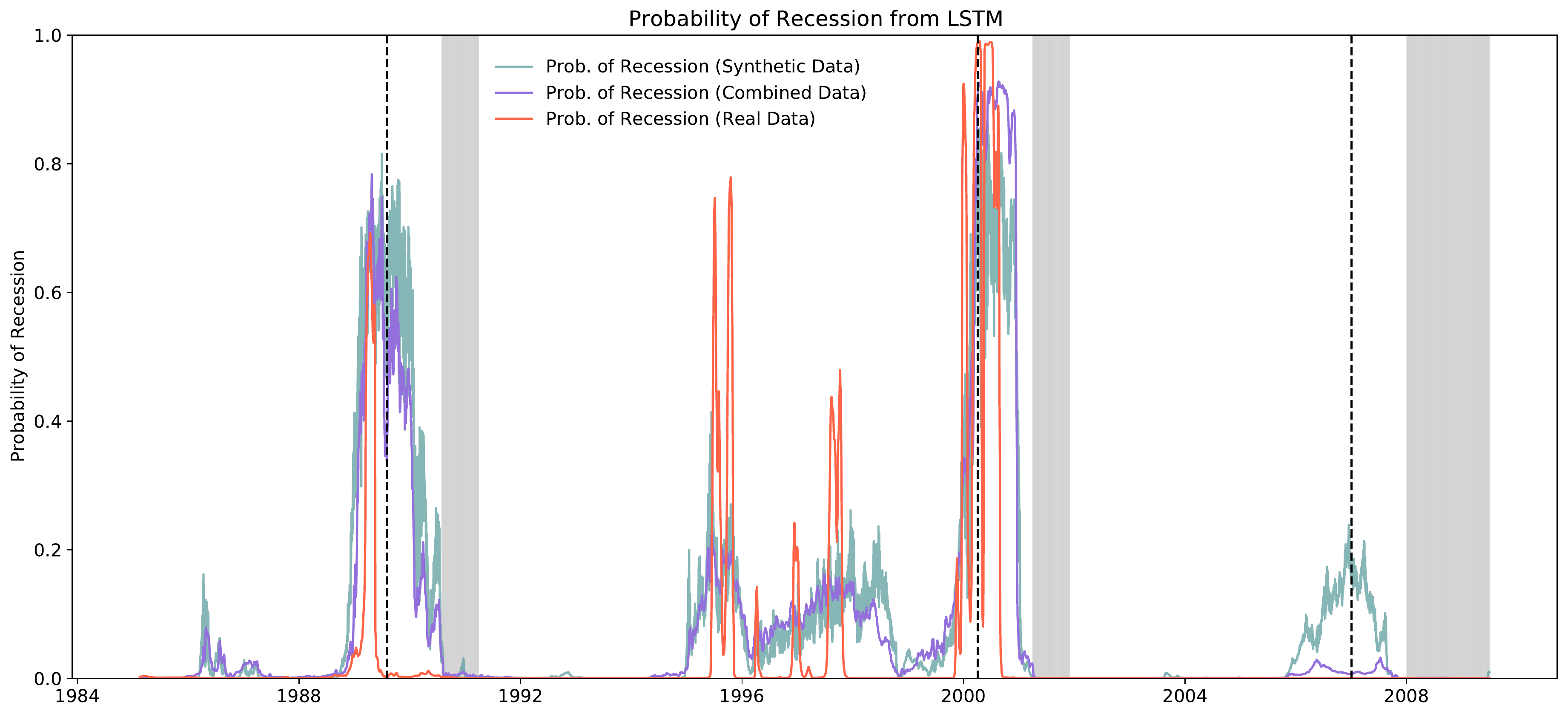}
    \captionsetup{width=.75\linewidth}
    \caption{
    Probability of Recession as predicted by LSTM model \\
    Note: Grey areas indicate recession. Black dashed lines indicate 250 days prior to the start of the next recession. \\
    Source: NBER, author's calculations
    }
    \label{fig:fig9}
\end{figure}

The probability of recession predictions produced by each LSTM classifier model can be found in Figure \ref{fig:fig9}. Here, the benefits of synthetic data are much more clear. The model trained on only real data gives short-lived spikes in probability and produces strong false positives in the mid to late 1990s. It also gives no signal at all of the 2008 recession. The model trained on synthetic data provides more prolonged signals of recession and produces weaker false positive signals in the mid to late 1990s. Again, the synthetic data produces much noisier estimates. The model trained on the combined data largely follows the same patterns of the model trained on synthetic data with somewhat less noise. Both the synthetic data and combined data models provide weak signals leading up to the 2008 recession. 

The LSTM models trained with synthetic data appear to perform quite well at providing high probabilities just prior to recessions while not being overly sensitive to minor yield curve inversions. The LSTM models overall did a worse job of predicting the 2008 recession relative to the logistic regression models, but within the class of LSTM models, synthetic data clearly improved the ability to predict the 2008 recession.

\begin{figure}
    \centering
    \includegraphics[scale = .45]{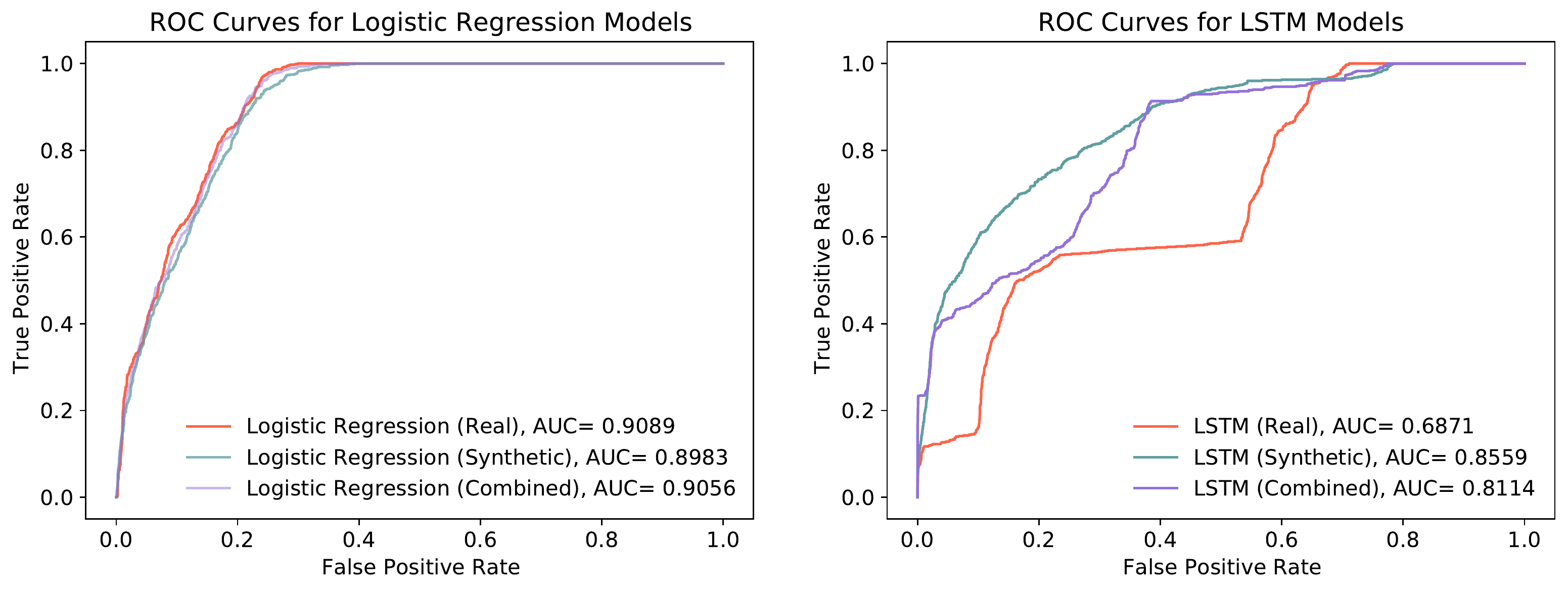}
    \captionsetup{width=.75\linewidth}
    \caption{
    ROC curves for all probability of recession models \\ 
    Source: author's calculations
    }
    \label{fig:fig10}
\end{figure}

The ROC curves for the LSTM models can also be found in Figure \ref{fig:fig10}. Here the AUC is clearly much higher for the model trained on synthetic data (0.86) and the model trained on combined data (0.81) than the AUC for the model trained only on real data (0.69). It appears that allowing the LSTM models to train on additional fake recessions does produce a significant improvement in predictive ability.

\section{Conclusion}
GANs have shown great promise in generating synthetic data for a wide range of tasks. This paper serves as a proof of concept, showing that GAN-generated synthetic time series data can be useful in forecasting applications, particularly in economic settings. For US financial and economic data, there is only one history. Previously, economic forecasters had access only to the information provided by the limited recessions that have occurred and the single history of Treasury yields. With GANs, and particularly DoppelGANger, researchers and forecasters can now produce as many synthetic histories as they please. Models can be trained on a much larger number of recessions and learn from a much wider range of samples. 

This paper shows that although these synthetic recessions are merely replicating the conditions of the true recessions they were trained on, they can actually provide information above and beyond the true recessions when performing downstream predictive tasks.  

The models used in this paper for economic forecasting and recession prediction are fairly simple models that rely only on 1-year and 10-year Treasury yields.  They are not fine-tuned or conditioned on other economic variables. They are merely illustrations of the comparative performance of simple models trained with real and synthetic data. This means that predictive performance could likely be vastly improved in future work by applying the synthetic data framework of this paper to more advanced econometric techniques, though that remains to be seen. Also note that despite the difficulty in capturing long-term autocorrelations for Treasury yield data, many of the models trained with synthetic data were still able to show improvements over models trained with only real data. Further improving the fidelity of the synthetic data could potentially produce even greater improvements.

In future work, this framework could be extended to a wide range of applications in which rare events need to be predicted - such as rare atmospheric events or the occurrence of disease - provided that high-fidelity synthetic data can be produced. This is certainly not the first paper to use synthetic data to improve predictive models, but it shows another useful application of the technique. As GANs and other machine learning techniques advance at a rapid pace, there is great potential for further exploration of how high-fidelity synthetic data can improve time series forecasting. 

\begin{appendices}
\section{RNNs and LSTM}
\label{appendix:A}
This appendix is adapted from Chapter 10 of \citet{Goodfellow-et-al-2016}.

Recurrent neural networks (RNNs) are a form of neural network designed to capture information from sequence data. The input data to RNNs are read sequentially, with information about previous inputs being stored as the RNN works its way through the sequence. The current state, $\bm{h}^{(t)}$, is a function of all previous states processed before step $t$ and current features, $\bm{x}^{(t)}$. 

$$\bm{h}^{(t)} = f(\bm{h}^{(t-1)}, \bm{x}^{(t)}; \bm{\theta})$$

Where $\bm{\theta}$ represents the parameters (weights) that the learning algorithm attempts to optimize. At each time step, the current state summarizes useful information from the current input, as well as all previous inputs, and then uses that summary as an input to the next time step. The advantage of using recurrence in this way is that the model can learn temporal patterns in the sequence of inputs. Using the same $f$ and $\bm{\theta}$ at each time step simplifies the training process to learning a single model for all time steps. An unrolled graph displaying an example RNN can be found in Figure \ref{fig:figA1}. 

\setcounter{figure}{0}
\counterwithin{figure}{section}
\begin{figure}[ht]
    \centering
    \includegraphics[scale = .45]{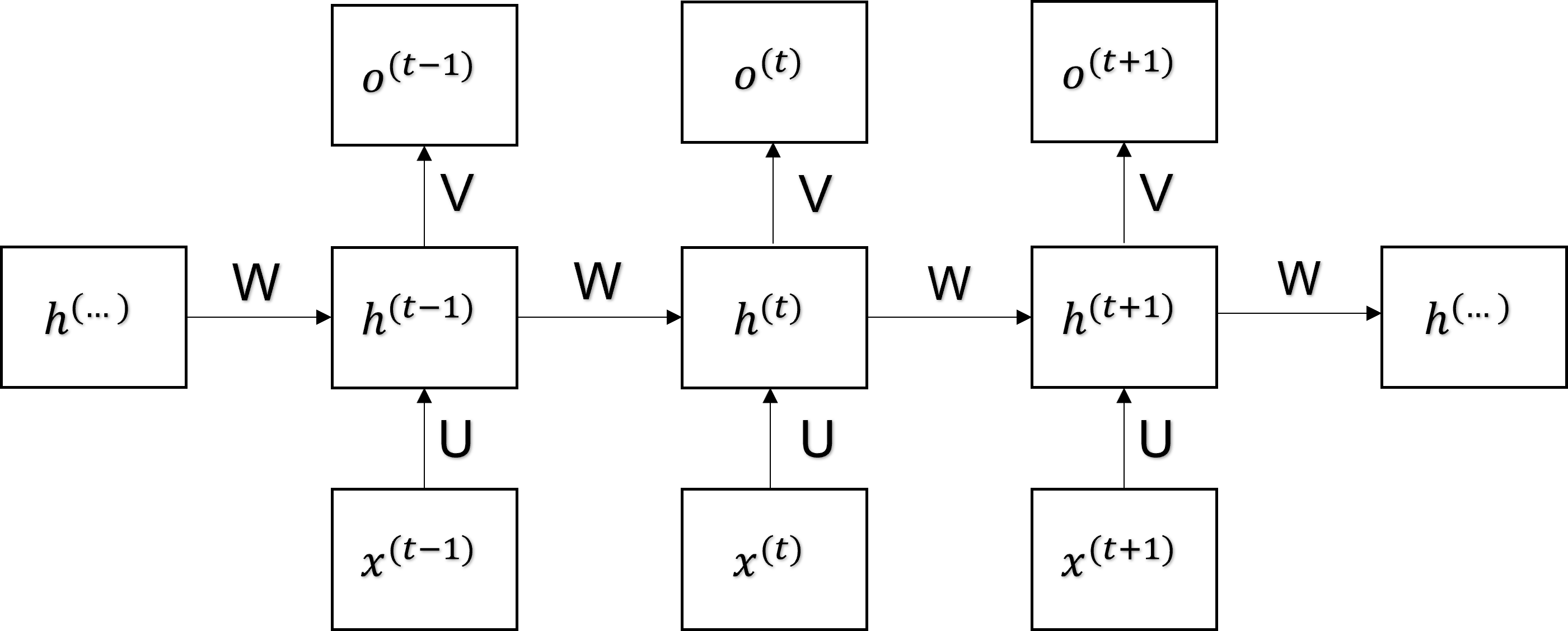}
    \captionsetup{width=.75\linewidth}
    \caption{
    An example of an unfolded RNN graph \\ 
    Source: Author, based on Figure 10.3 of \citep{Goodfellow-et-al-2016} \\
    Note: $\bm U, V, W$ represent weights between layers. Not shown are activation functions that will accompany these weights. 
    }
    \label{fig:figA1}
\end{figure}

In Figure \ref{fig:figA1}, $\bm{h^{(i)}}$ represents hidden states that are used as an input to the next time step, $\bm{x^{(i)}}$ represents the features input at each time step, and $\bm{o^{(i)}}$ represents the outputs of each time step. $W$ represents the weights that parameterize connections between hidden layers, $U$ represents weights between input layers and hidden layers, and $V$ represents weights between hidden layers and output layers. $\bm{o}^{(t)}$ will be compared to the target $\bm{y}^{(t)}$ when computing the loss function for backpropagation. There are many different configurations of RNNS, and this chart represents one of the most simple configurations. 

Long-term interactions present in sequences can be hard to estimate with a standard RNN. As the gradients are backpropagated through a large number of time steps, they tend to become smaller and smaller until they vanish. It is also possible to create exploding gradients from exponential growth. In the more common case of vanishing gradients, time steps further back in the sequence will have very small gradients, thus making it difficult to learn long-term interactions.

\begin{figure}[ht]
    \centering
    \includegraphics[scale = .45]{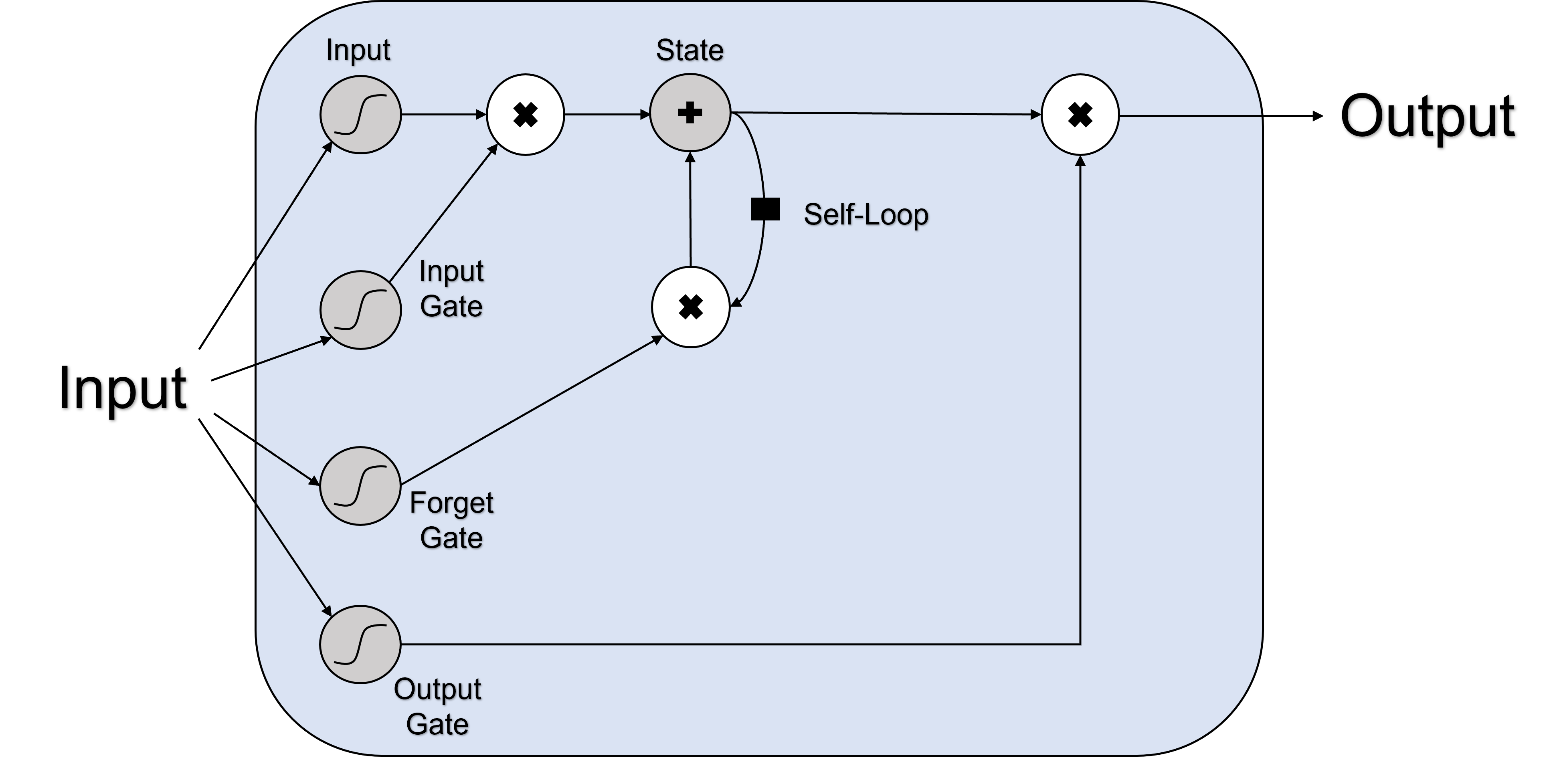}
    \captionsetup{width=.75\linewidth}
    \caption{
    A single LSTM cell. Cells replace the traditional hidden units of an RNN. \\ 
    Source: Author, based on Figure 10.16 of \citep{Goodfellow-et-al-2016} \\
    Note: Gates use a sigmoid activation function.
    }
    \label{fig:figA2}
\end{figure}

A popular way of dealing with the vanishing gradient problem is by using a gated RNN such as "long short-term memory" (LSTM) networks. The name long short-term memory refers to the fact that these networks capture the short-term changes in state from one step to the next (short-term memory), but also sometimes choose to keep past information for many steps (long-term memory) \citep{Glassner}. LSTM networks replace the hidden layers of the common RNN with LSTM "cells". These cells are composed of multiple networks. A network known as the "forget gate" controls how much past information to discard by moving weights on previous units toward zero. Another network known as "input gate" controls whether inputs should be accumulated into the current state. Finally, an "output gate" controls the cell's output. An LSTM cell can be seen in Figure \ref{fig:figA2} 

\counterwithin{figure}{section}

The diagram of an LSTM cell displays the basic function of gates within the LSTM cell. The inputs ($\bm{x^{(t)}}$ and $\bm{h^{(t-1)}}$) are provided to the input unit and each of the three gates. All of the gates use a sigmoid activation function. The input gate can "shut off" the input through a sigmoid function and thus determines what inputs to store in the current state. The state has a recurrent connection to itself (a self-loop) whose effect is controlled by the output of the forget gate. This effectively decides when it is appropriate to "forget" prior information. The output also has a gate with a sigmoid activation. Each gate is a network with its own input and recurrent weights to train. All of these gates allow for the LSTM network to dynamically control the amount of long-term memory that is relevant. Most importantly, this framework allows the gradients to backpropagate for long durations without vanishing or exploding.

\section{Synthetic Data fidelity metrics for classification models}
\label{appendix:B}

The figures below display the fidelity metrics for the synthetic data generated for the recession classifier task. The correlation between 1-year and 10-year Treasury yields in the training samples was 0.950. The same correlation in the synthetic samples was 0.953.

\setcounter{figure}{0}
\counterwithin{figure}{section}
\begin{figure}[ht]
    \centering
    \includegraphics[scale = .3]{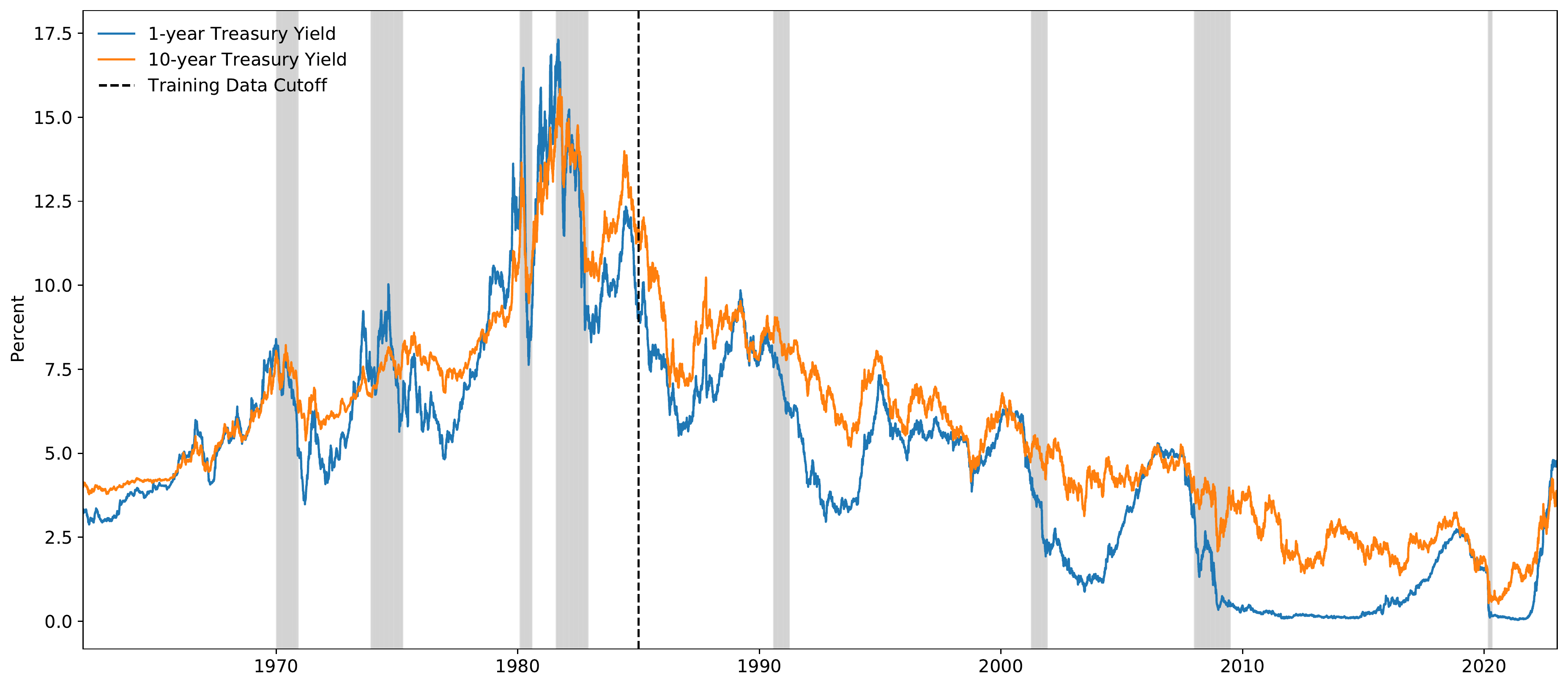}
    \captionsetup{width=.75\linewidth}
    \caption{
    U.S. Treasuries Yield Data\\
    Note: Grey areas indicate recession. The training sample contained 4 recessions, leaving 3 recessions for testing the classification models (the 2020 recession is excluded from the test set). \\
    Sources: Board of Governors of the Federal Reserve System, Federal Reserve Bank of St. Louis, NBER
    }
    \label{fig:figB1}
\end{figure}

\begin{figure}[ht]
    \centering
    \includegraphics[scale = .4]{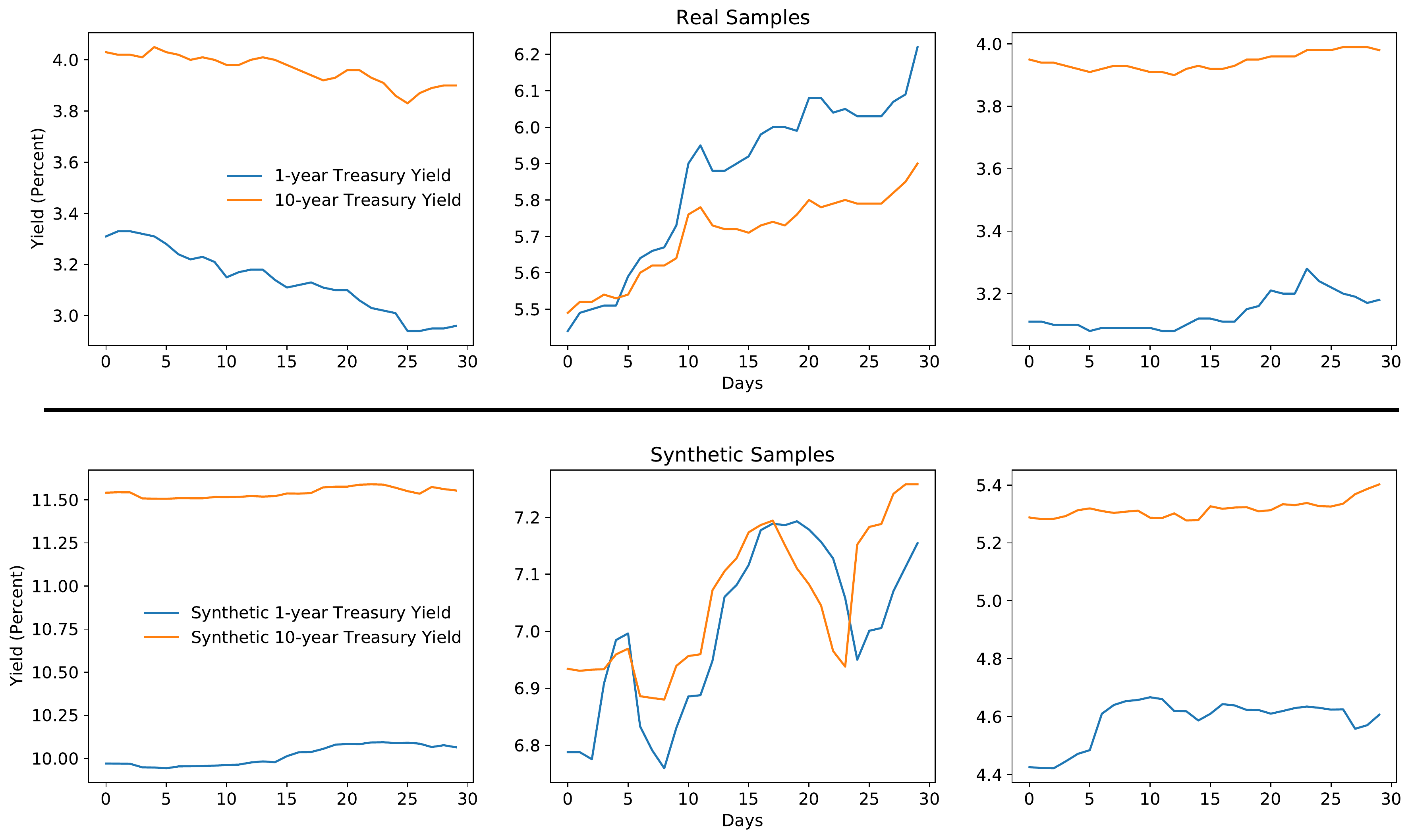}
    \captionsetup{width=.75\linewidth}
    \caption{
    Comparison of real samples to synthetic samples for classifier \\
    Note: The top row contains true data samples from the training set.  The bottom row contains generated data from DoppelGANger.  \\
    Sources: Board of Governors of the Federal Reserve System, Federal Reserve Bank of St. Louis, NBER, author
    }
    \label{fig:figB2}
\end{figure}

\begin{figure}[ht]
    \centering
    \includegraphics[scale = .5]{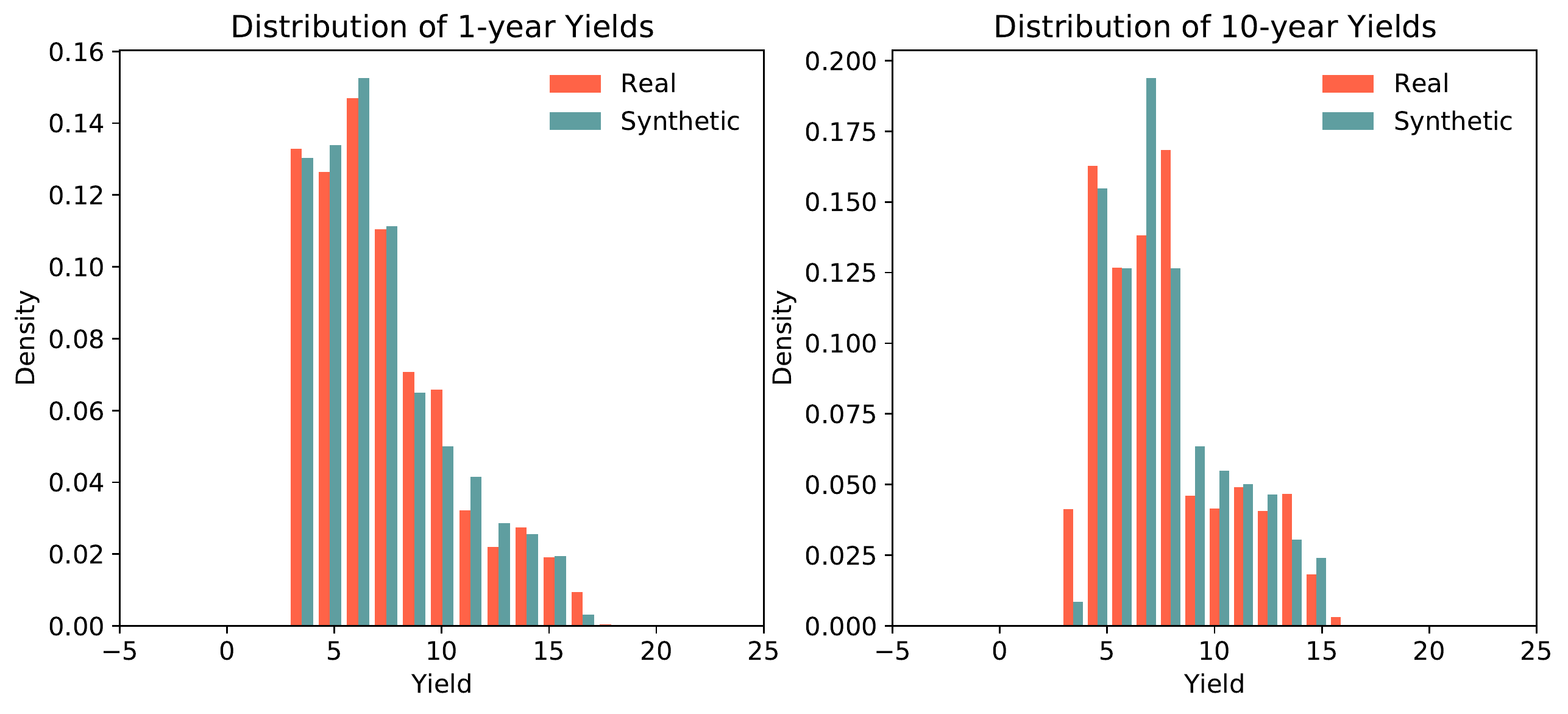}
    \captionsetup{width=.75\linewidth}
    \caption{
    Distributions of real and generated yields for classifier \\
    Source: author's calculations
    }
    \label{fig:figB3}
\end{figure}

\begin{figure}[ht]
    \centering
    \includegraphics[scale = .5]{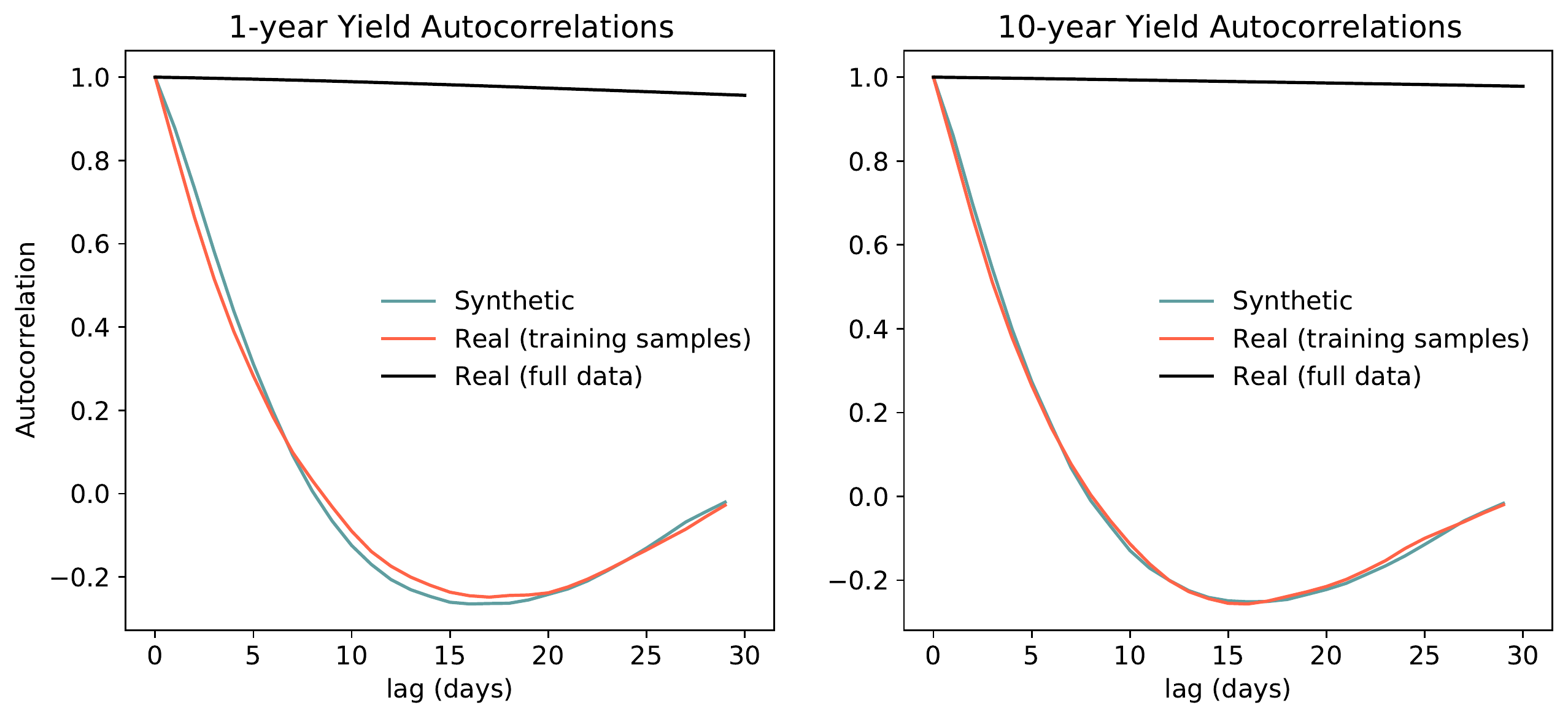}
    \captionsetup{width=.75\linewidth}
    \caption{
    Estimated autocorrelations for classifier \\
    Note: The synthetic and real (training samples) values are estimated as an average of autocorrelations across many samples. The real (full data) autocorrelation is estimated on the full range of training data before being split into training samples. \\ 
    Source: author's calculations
    }
    \label{fig:figB4}
\end{figure}

\clearpage

\section{LSTM Model Diagrams}
\label{appendix:C}

\setcounter{figure}{0}
\counterwithin{figure}{section}
\begin{figure}[ht]
    \hspace*{-2cm}
    \centering
    \includegraphics[scale = .45]{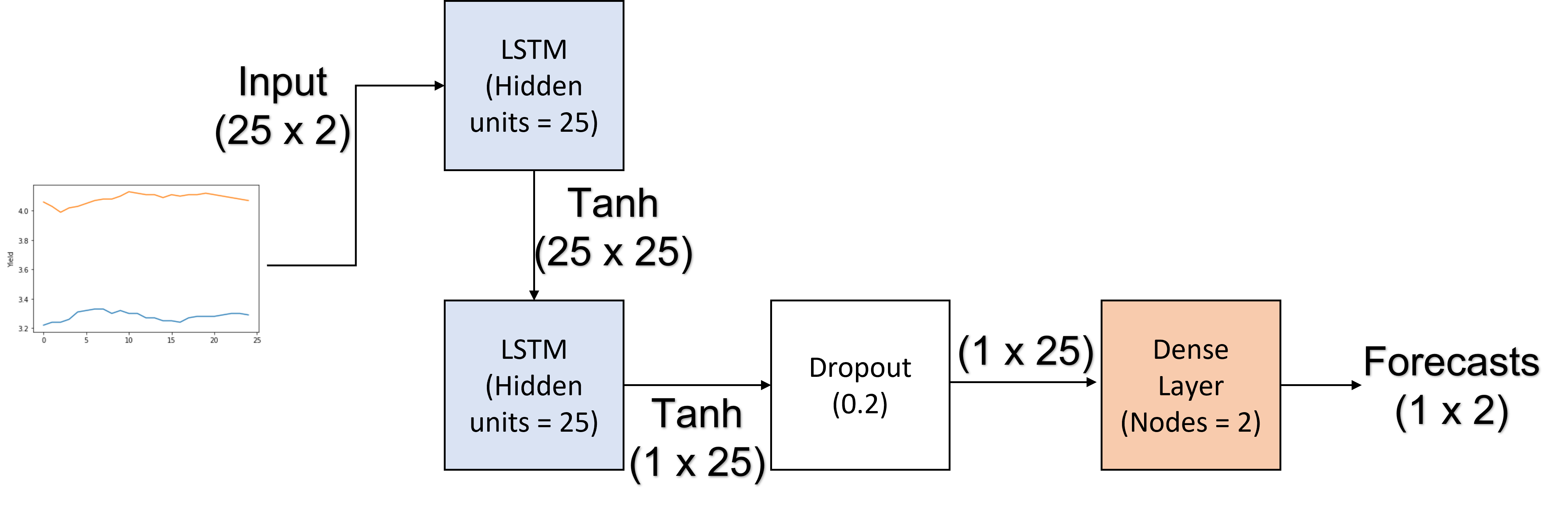}
    \captionsetup{width=.75\linewidth}
    \caption{
    Diagram of network used for forecasting task\\
    Note: This diagram represents the LSTM model used for 1-day ahead forecasts. For 15-day ahead forecasts, the model was altered to allow for an output of shape 15x2. \\
    Source: Author
    }
    \label{fig:figC1}
\end{figure}

\vspace{2cm}

\begin{figure}[ht]
    \hspace*{-2cm}
    \centering
    \includegraphics[scale = .45]{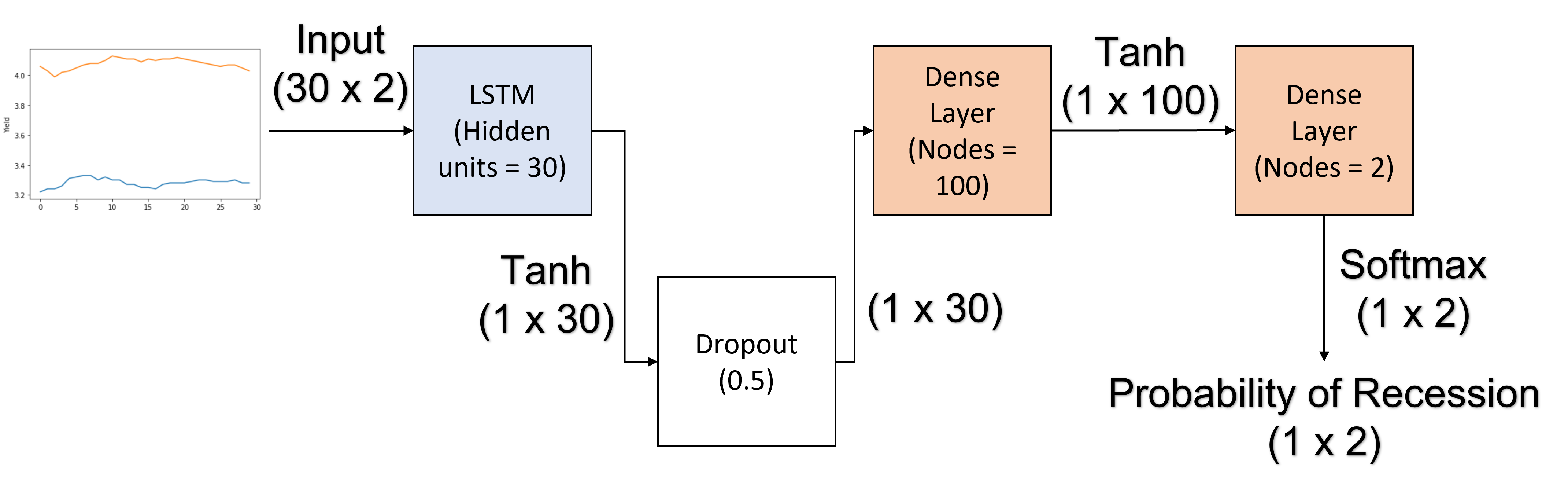}
    \captionsetup{width=.75\linewidth}
    \caption{
    Diagram of network used for classification task\\
    Source: Author
    }
    \label{fig:figC2}
\end{figure}

\end{appendices}

\clearpage

\bibliographystyle{apacite}
\bibliography{references}

\end{document}